\def\year{2022}\relax
\documentclass[letterpaper]{article} % DO NOT CHANGE THIS
\usepackage{aaai22}  % DO NOT CHANGE THIS
\usepackage{times}  % DO NOT CHANGE THIS
\usepackage{helvet}  % DO NOT CHANGE THIS
\usepackage{courier}  % DO NOT CHANGE THIS
\usepackage[hyphens]{url}  % DO NOT CHANGE THIS
\usepackage{graphicx} % DO NOT CHANGE THIS
\usepackage{natbib}  % DO NOT CHANGE THIS AND DO NOT ADD ANY OPTIONS TO IT
\usepackage{caption} % DO NOT CHANGE THIS AND DO NOT ADD ANY OPTIONS TO IT
\DeclareCaptionStyle{ruled}{labelfont=normalfont,labelsep=colon,strut=off} % DO NOT CHANGE THIS
\usepackage{algorithm}
\usepackage{algorithmic}
\usepackage{newfloat}
\usepackage{listings}
\floatstyle{ruled}
\newfloat{listing}{tb}{lst}{}
\floatname{listing}{Listing}
\usepackage{tikz-cd}
\usetikzlibrary{decorations.pathmorphing}
\usepackage{mathtools}
\usepackage{tabularx}
\usepackage{multirow}
\usepackage{amsmath}
\usepackage{amssymb}
\usepackage{booktabs}

\makeatletter
\newcommand\makebig[2]{%
	\@xp\newcommand\@xp*\csname#1\endcsname{\bBigg@{#2}}%
	\@xp\newcommand\@xp*\csname#1l\endcsname{\@xp\mathopen\csname#1\endcsname}%
	\@xp\newcommand\@xp*\csname#1r\endcsname{\@xp\mathclose\csname#1\endcsname}%
}
\makeatother
\makebig{biggg} {3.0}
\makebig{Biggg} {3.5}
\makebig{bigggg}{4.0}
\makebig{Bigggg}{4.5}
\title{ Optimizing Edge Detection for Image Segmentation with Multicut Penalties }
\author{
    Steffen Jung, \textsuperscript{\rm 1}
    Sebastian Ziegler, \textsuperscript{\rm 2}
    Amirhossein Kardoost, \textsuperscript{\rm 2}
    Margret Keuper \textsuperscript{\rm 1,3}
    \\
}
\affiliations {
    \textsuperscript{\rm 1} Max Planck Institute for Informatics, Saarland Informatics Campus, \\
    \textsuperscript{\rm 2} University of Mannheim, 
    \textsuperscript{\rm 3} University of Siegen \\
    steffen.jung@mpi-inf.mpg.de,
    sziegler@mail.uni-mannheim.de,
    kardoostamirhossein@gmail.com,
    margret.keuper@uni-siegen.de
}

\usepackage{bibentry}

\begin{document}

\maketitle
\begin{abstract}
The Minimum Cost Multicut Problem (MP) is a popular way for obtaining a graph decomposition by optimizing binary edge labels over edge costs.
While the formulation of a MP from independently estimated costs per edge is highly flexible and intuitive, solving the MP is NP-hard and time-expensive.
As a remedy, recent work proposed to predict edge probabilities with awareness to potential conflicts by incorporating cycle constraints in the prediction process.
We argue that such formulation, while providing a first step towards end-to-end learnable edge weights, is suboptimal, since it is built upon a loose relaxation of the MP.
We therefore propose an adaptive CRF that allows to progressively consider more violated constraints and, in consequence, to issue solutions with higher validity.
Experiments on the BSDS500 benchmark for natural image segmentation as well as on electron microscopic recordings show that our approach yields more precise edge detection and image segmentation.
\end{abstract}

\section{Introduction}

\begin{figure}[t!]
	\centering
	
	\begin{tabular}{@{}c@{ }c@{}}
	    \includegraphics[width=0.22\textwidth]{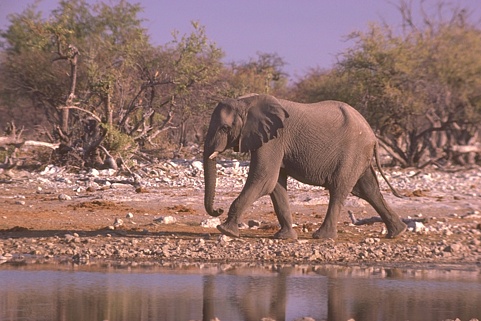} & 
		\includegraphics[width=0.22\textwidth]{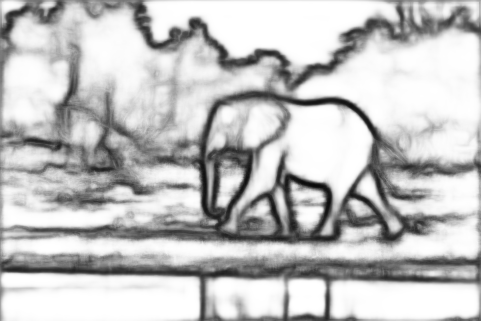} \\  
		(a) Original Image  & (b) RCF  \\[6pt]
		\includegraphics[width=0.22\textwidth]{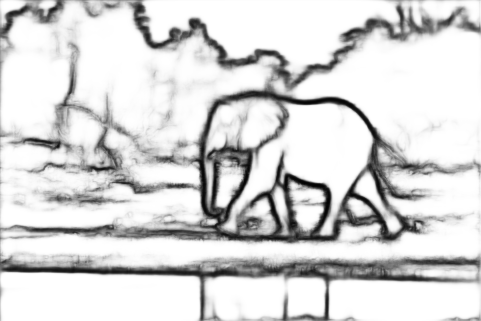} &
		\includegraphics[width=0.22\textwidth]{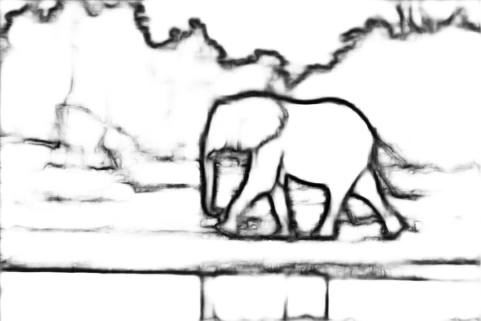}\\
		(c) RCF-basic-CRF (ours) & (d) RCF-adaptive-CRF (ours)  \\[6pt]
		
	\end{tabular}
	\caption{
	(a) Example BSDS500 \cite{bsds} test image,
	(b) the edge map produced with the RCF edge detector \cite{rcf-2019},
	(c) the edge map by RCF-CRF using the basic CRF by \citet{end2end},
	and (d) the edge map from RCF-CRF using our adaptive CRF.
	The adaptive CRF promotes closed contours and removes trailing edges.
	}
	\label{intro-fig}
\end{figure}

Image Segmentation, one of the fundamental problems in Computer Vision, is the task of partitioning an image into multiple disjoint components such that each component is a meaningful part of the image.
It is a low-level technique that assigns a label to every pixel in the image, and plays an important role in many areas like medical image analysis~\cite{xian2018automatic}, for example localizing tumors~\cite{litjens2017survey} or aneurysms \cite{de2004interactive}, and in other fields like satellite imagery and forensics \cite{ghosh2019understanding}.
Former successful image segmentation approaches are graph-based, i.e.,~they map image elements, for example pixels or superpixels, onto a graph.
While there are several different approaches to obtain a decomposition from a graph, the Minimum Cost Multicut Problem (MP)~\cite{chopra1993partition,deza1997geometry}, also called Correlation Clustering~\cite{cc}, is a well-known approach for image segmentation.
Here, the number of components is unknown beforehand and no bias is assumed in terms of component sizes.
The resulting segmentation is only determined by the input graph~\cite{keuper_lifted}, for which edge features can be generated by an edge detector such as Convolutional Neural Networks (CNNs) that predict edge probabilities (see Fig. \ref{intro-fig} for examples).

A feasible solution to the MP decomposes the graph into disjoint subgraphs via a \emph{binary} edge labeling.
The decomposition is enforced by cycle consistency constraints in general graphs.
If a path exists between two nodes where the direct edge between both is cut, then this constraint is violated.
\citet{end2end} introduced relaxed cycle constraints, i.e.~constraints evaluated on non-binary network predictions, as higher-order potentials in a Conditional Random Field (CRF) to allow for end-to-end training of graph-based human pose estimation.
By doing so, cycle constraint violations become a supervision signal and can be reduced when training the feature extractor and the CRF jointly.
We transfer this approach to image segmentation, where such constraints enforce that object contours are closed. Yet, we observe that optimizing non-binary network predictions instead of binary edge labels only leads to few additionally closed contours. This is consistent with prior works on linear program multicut solvers which show that relaxations of the cycle constraints to non-binary edge labels are too loose in practice \cite{kappes-2011}.  
In the CRF formulation, we propose to alleviate this issue by enforcing more binary (i.e. closer to $0$ or $1$) edge predictions that lead to less violated cycle constraints.
Our contributions are twofold.
We are the first to address boundary driven image segmentation using multicut constraints.
To this end, we combine a CRF with the neural edge detection model Richer Convolutional Features (RCF) by \citet{rcf-2019} to design an end-to-end trainable architecture that inputs the original image and produces a graph with edge probabilities optimized by the CRF.
Secondly, we propose an approach that progressively uses "closer to binary" boundary estimates in the optimization of the CRF model by \citet{end2end} and thus resolves progressively more boundary conflicts.
In consequence, the end-to-end trained network yields more and more certain predictions throughout the training process while reducing the number of violated constraints.
We show that this improved certainty yields improved results for edge detection and segmentation on the BSDS500 benchmark~\cite{bsds} and the ISBI 2012 neuronal structure segmentation~\cite{isbi}.
An example result is given in Fig. \ref{intro-fig}.
Compared to the plain RCF architecture as well as to the RCF with the CRF by~\citet{end2end}, our approach issues cleaner, less cluttered edge maps with closed contours.

\section{Related Work}
In the following, we first review the related work on edge detection and minimum cost multicuts in the context of image segmentation.

Edge detection in the context of image segmentation is usually based on learning-driven approaches that facilitate to learn to discriminate between object boundary and other sources of brightness change such as textures.
Structured random forests have been employed in~\citet{DollarICCV13edges} to train an edge detector on local image patches.
\citet{hed} proposed a CNN-based approach called holistically nested edge detection (HED) that intrinsincally leverages multiple edge map resolutions.
Similarly,~\citet{Kokkinos_2017_CVPR} propose an end-to-end CNN for low-, mid- and high-level vision tasks such as boundary detection, semantic segmentation, region proposal generation, and object detection in a single network based on multi-scale learning.
Convolutional oriented boundaries (COB)~\cite{cob} compute multiscale oriented contours and region hierarchies in a single forward pass and provide boundary orientation estimates.
Such boundary orientations are needed as input along with the edge maps in order to compute hierarchical image segmentations in frameworks such as MCG \cite{APBMM2014,PABMM2015} or gpb-owt-ucm~\cite{bsds}.
If not estimated by the network, they have to be approximated, for example using filter-based approaches \cite{DollarICCV13edges}.
In~\citet{bdcn}, a bi-directional cascade network (BDCN) is proposed for edge detection of objects in different scales, where individual layers of a CNN model are trained by labeled edges at a specific scale.
Similarly, to address the edge detection in multiple scales and aspect ratios,~\citet{rcf-2019} provide an edge detector using richer convolutional features (RCF) by exploiting multiscale and multilevel information of objects.
Although BDCN provides slightly better edge detection accuracy on the BSDS dataset~\cite{bsds}, we base our approach on the RCF edge detection framework because of its more generic training procedure.

The multicut approach has been extensively used for image segmentation, for example  in~\citet{kappes-2015-ssvm,kappes-2011,keuper_lifted,bsds,beier2015fusion,andres-2011}.
Due to the NP-hardness of the MP, segmentation has often been addressed on pre-defined superpixels~\cite{andres-2013,kappes-2011,kappes-2013-arxiv,beier2015fusion}. While \citet{kappes-2015-ssvm} utilize multicuts as a method for discretizing a grid graph defined on the image pixels, where the local connectivity of the edges define the join/cut decisions and the nodes represent the image pixels, \citet{keuper_lifted} proposed long-range terms in the objective function of the multicut problem defined on the pixel grid.
Such terms are efficient to deal with image decomposition problems where there is no clear cut or join signal along blurry edges. 
An iterative fusion algorithm for the MP has been proposed in \citet{beier2015fusion} to decompose the graph. 
\citet{andres-2011} propose a graphical model for probabilistic image segmentation and globally optimal inference on the objective function with higher orders.
A similar higher order approach is proposed also in~\citet{kappes-2013-arxiv,kim-2014} for image segmentation.

\section{End-to-end Learning of Edge Weights for Graph Decompositions}
\subsection{Cycle constraints in the Multicut Problem}

\begin{figure}%[H]
    \centering
    \includegraphics[width=0.9\linewidth]{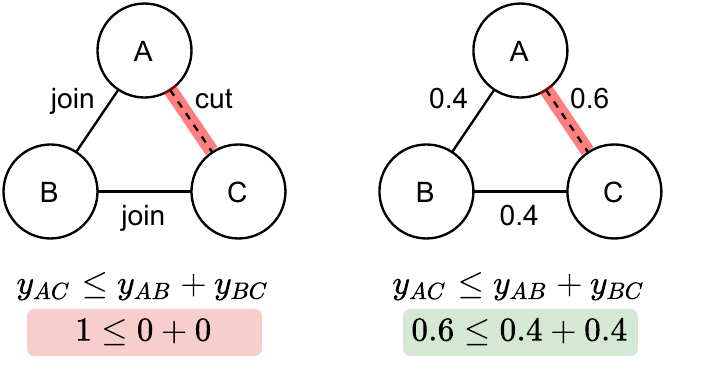}
	\caption[
	Invalid Solution to the Multicut Problem]{Invalid solution to the Multicut Problem.
	(left) Covered by cycle inequality constraints given the integer solution.
	(right) Not covered given a relaxed solution that results in (left) when rounded.}
	\label{invalid2}
	
\end{figure}

The MP is based on a graph $ G = (V,E) $, where every pixel (or superpixel) is represented by an individual node or vertex $ v\in V $. Edges $e\in E $ encode whether two pixels belong to the same component or not.
The goal is to assign every node to a cluster by labeling the edges between all nodes as either "cut" or "join" in an optimal way based on edge costs $c_e$.
One of the main advantages of this approach is that the number of components is not fixed beforehand, contrary to other clustering algorithms, and is determined by the input graph instead.
Since the number of segments in an image cannot be foreseen, the MP is a well-suited approach.
The MP can be formulated as integer linear program~\cite{chopra1993partition,deza1997geometry} 
with objective function $ c:E\rightarrow\mathbb{R} $ as follows:
\begin{equation}
\label{mp_objective}
\min_{y\in\left\{0, 1\right\}^{E}} \sum \limits_{e\in E}c_{e}y_{e}
\end{equation}
subject to 
\begin{equation}
\label{cycle_constraints}
\forall C \in cc(G)\forall e \in C: y_{e}\leq \sum \limits_{e'\in C \setminus\left\{e\right\} }y_{e'},
\end{equation}
where $y_e$ is the binary labeling of an edge $e$ that can be either $0$ (join) or $1$ (cut),
and $cc(G)$ represents the set of all chordless cycles in the graph.
If the cycle inequality constraint in Eq.~\eqref{cycle_constraints} is satisfied, the MP solution results in a decomposition of the graph and therefore in a segmentation of the image.
Informally, cycle inequality constraints ensure that there cannot be exactly one cut edge in any chordless cycle.
However, computing an exact solution is not tractable in many cases due to the NP-hardness of the problem.
Relaxing the integrality constraints such that $y \in [0,1]^{E}$ can improve tractability, however, valid edge label configurations are not guaranteed in this case.
An example can be seen in Fig. \ref{invalid2}, where node $B$ is supposed to be in the same component as $A$ and $C$, however, $A$ and $C$ are considered being in different components.
Infeasible solutions have to be repaired in order to obtain a meaningful segmentation of the input image.
This can be achieved by using heuristics like in \citet{beier2014cut, kardoost2018solving, keuper_lifted, pape2017solving}.

\subsection{Incorporating cycle constraints into a CRF}
To improve the validity of relaxed solutions, \citet{end2end} reformulate the MP as an end-to-end trainable CRF based on a formulation as Recurrent Neural Network (RNN) by \citet{crfasrnn}.
By doing so, they are able to impose costs for violations of cycle inequality constraints during training.
This is accomplished by first transforming the MP into a binary cubic problem, considering all triangles in the graph:
\begin{equation}
\label{costs}
\min_{y\in\left\{0, 1\right\}^{E}} \sum \limits_{e\in E}c_{e}y_{e}+\gamma \sum \limits_{\left\{u, v, w\right\}\in\begin{pmatrix} V\\3 \end{pmatrix}}\begin{aligned}(y_{uv}\overline{y}_{vw}\overline{y}_{uw}+\\\overline{y}_{uv}y_{vw}\overline{y}_{uw}+\\\shoveleft{\overline{y}_{uv}\overline{y}_{vw}y_{uw}).\:\,}\end{aligned}
\end{equation}
This formulation moves cycle inequalities into the objective function by incurring a large cost $\gamma$ whenever there is an invalid edge label configuration like $(cut,join,join)$ in a clique (as shown in Fig. \ref{invalid2} -- all other orders implied).
The binary cubic problem is then transformed to a CRF by defining a random field over the edge variables $y = (y_{1},y_{2},...,y_{|E|})$, which is conditioned on the image $I$. 
The cycle inequality constraints are incorporated in the form of higher-order potentials as they always consider three edge variables.
The combination of unary and third-order potentials yields the following energy function building the CRF:
\begin{equation}
\label{crf-potentials}
E(y|I)=\sum \limits_{i}\psi^{U}_{i}(y_{i})+\sum \limits_{c}\psi^{Cycle}_{c}(y_{c})
\end{equation}
The energy $ E(y|I) $
is the sum of the unary potentials and the higher-order potentials.
For the latter \cite{end2end} used pattern-based potentials proposed by \citet{komodakis2009beyond}: 
\begin{equation}
\psi^{Cycle}_{c}(y_{c}) = \begin{cases} \gamma_{y_{c}} & \mathrm{if} \ y_{c}\in P_{c} \\ \gamma_{max} & \mathrm{otherwise} \end{cases} 
\end{equation}
$P_{c}$ represents the set of all valid edge label configurations, which are $(join,join,join)$, $(join,cut,cut)$, and $(cut,cut,cut)$.
The invalid edge label configuration is $(cut,join,join)$.
The potential assigns a high cost $\gamma_{max}$ to an invalid labeling and a low cost $\gamma_{y_{c}}$ to a valid labeling.

Such CRFs can be made end-to-end trainable
using Mean-Field Updates, as proposed by \citet{crfasrnn}. 
This approach computes an auxiliary distribution $Q$ over $y$
such as to minimize the Kullback-Leibler Divergence (KL-Divergence) \cite{csiszar1975divergence, csiszar2007entropy} between $Q$ and the true posterior distribution of $y$.
This step of optimizing $Q(y_i)$ instead of $y_i$ can be interpreted as relaxation.
Instead of considering violated constraints on binary edge variables (see Eq. \eqref{mp_objective}), we optimize probabilities of $y_i$ taking a certain label $l$ and optimize $Q(y_i=l)$ which admits values in the interval $[0,1]$.

\citet{crfasrnn} reformulate the update steps as individual CNN layers and then repeat this stack multiple times in order to compute multiple mean-field iterations.
The repetition of the CNN layer stack is treated equally to an RNN and remains fully trainable via backpropagation.
\citet{vineet2014filter} extends this idea by incorporating higher-order potentials in the form of pattern-based potentials and co-occurrence potentials.
For the CRF by \citet{end2end} the corresponding update rule becomes:
\begin{multline}
\label{mean-field-update}
Q^{t}_{i}(y_{i}=l)= \\
\frac{1}{Z_{i}}exp \bigggl \{
-\sum \limits_{c\in C} \bigggl (
\overbrace{\sum \limits_{p\in P_{c|y_{i}=l}}\biggl (\prod \limits_{j\in c,j\neq i}Q^{t-1}_{j}(y_{j}=p_{j})\biggr )}^\text{valid labeling case gets low costs}\gamma_{p}
\\+ \gamma_{max}\underbrace{ \biggl( 1- \biggl( \sum \limits_{p\in P_{c|y_{i}=l}}\biggl (\prod \limits_{j\in c,j\neq i}Q^{t-1}_{j}(y_{j}=p_{j})\biggr )
	\biggr )
	\biggr )}_\text{inverse of the valid labeling case gets high costs}
\bigggr )
\bigggr \},
\end{multline}
where $Q^{t}_{i}(y_{i}=l)\in [0,1]$ is the $Q$ value of one edge variable at mean-field iteration $t$
with fixed edge label $l$ (either $cut$ or $join$).
$P_{c|y_{i}=l}$ represents the set of valid edge label configurations according to Eq.~\eqref{cycle_constraints}, where the considered edge label $l$ is fixed.
Looking at the case where $y_{i}=1$ ($cut$), possible valid configurations are ($y_{i}$,$0$,$1$), ($y_{i}$,$1$,$0$), and ($y_{i}$,$1$,$1$).
For all valid labelings the other two variables $y_{j\neq i}$ in clique $c$ are taken into account. Their multiplied label probabilities from iteration $t$--1 for every valid label set, are summed and then multiplied with the cost for valid labelings $ \gamma_{p} $. The inverse of the previous result is then multiplied with $ \gamma_{max} $. Taking the inverse is equal to computing the same as in the valid case but with all possible invalid labelings for the considered edge variable. The results of all cliques $C$, which the variable $y_{i}$ is part of, are summed. The updated $Q(y_i=l)$ are projected onto $[0,1]$ using softmax.
Costs $ \gamma_{p} $ and $ \gamma_{max} $ are considered as trainable parameters, and the update rule is differentiable with regard to the input $Q(y_i=l)$ (a network estimate of the likelihood of $y_i$ taking label $l$) and the cost parameters.
Since non-binary values for $Q(y_i=l)$ are optimized, invalid configurations are assigned lower costs when they become uncertain, i.e., as $Q(y_i=l)\rightarrow 0.5$.

\subsection{Cooling Mean-Field Updates}

There have been various attempts tightening the relaxation of the MP.
For example \citet{swoboda2017message} incorporate odd-wheel inequalities and \citet{kappes-2013-arxiv} use additional terminal cycle constraints.
While these approaches can achieve tighter solution bounds, they involve constraints defined on a larger number of edges.
A formulation of such constraints in the context of higher-order CRFs, requiring at least an order of 4, is intractable.
Therefore, we choose an alternative, more straight forward approach -- we push the network predictions $Q(y)$, which we interpret as relaxed edge labels, progressively closer to $0$ or $1$ in the mean-field update by introducing a cooling scheme.
This not only issues solutions closer to the integer solution, but also provides a better training signal to the CRF, where cycle constraints penalize non-valid solutions more consistently.
For this, we substitute $Q^{t-1}_{i}(y_{i}=p_j)$ in Eq.~\eqref{mean-field-update} by $\phi\left(Q^{t-1}_{j}(y_{j}=p_{j}),k\right)$,
where $\phi$ is the newly introduced function modifying a probability $q$ as follows:
\begin{equation}
\label{eq:phi}
\phi(q,k) = \begin{cases} 1-(1-q)^{k} & \mathrm{if} \ q>=0.5 \\ q^{k} & \mathrm{otherwise.} \end{cases} 
\end{equation}
Here, $k$ is an exponent that can be adapted during training, and $q$ is the edge probability.
This function pushes values larger or equal to $0.5$ closer to $1$ and values below $0.5$ closer to $0$.
Using this transformation in the mean-field iterations reduces the intregrality gap
and therefore enforces the MP cycle constraints more strictly.
This effect is amplified the larger the exponent $k$ becomes.

Choosing a large $k$ from the start hampers network training as the edge detection parameters would need to adapt too drastically for close-to-binary edges.
We thus aim at adapting $k$ throughout the training process such as to first penalize violated cycle constraints on the most confident network predictions and progressive addressing violated constraints on less confident (i.e.~"less binary") estimates.
We therefore propose a cooling schedule, which defines criteria upon which the parameter $k$ is increased, as follows:
\begin{equation}
k = \begin{cases} k+0.05 & \mathrm{if} \ N(C_{inv})<a \\ k & \mathrm{otherwise,} \end{cases} 
\label{eq:cool}
\end{equation}
where $N(C_{inv})$ is the average number of invalid cycles (that do not adhere to the cycle inequality constraints) across all images and the hyperparameter $a$ is the number of allowed cycle constraint violations before increasing the exponent $k$. The exponent therefore is increased after every epoch in which the number of invalid (relaxed) cycles has been decreased below the threshold~$a$. 

Leveraging the mean-field update for image segmentation requires an edge detection network that provides values for the CRF potentials.
A possible learning-based approach that provides high quality edge estimates is the RCF~\cite{rcf-2019} architecture.
In practice, the edge detection network is first pre-trained until convergence, and then fine-tuned with the CRF.
Parameter $k$ is initialized as $1$ and updated using Eq.~\eqref{eq:cool} during training. 

\section{Richer Convolutional Features - CRF}

\begin{figure*}[t]
    \centering

    \begin{tabular}{cc}
        \multirow{8}{*}[-4em]{\includegraphics[height=9cm]{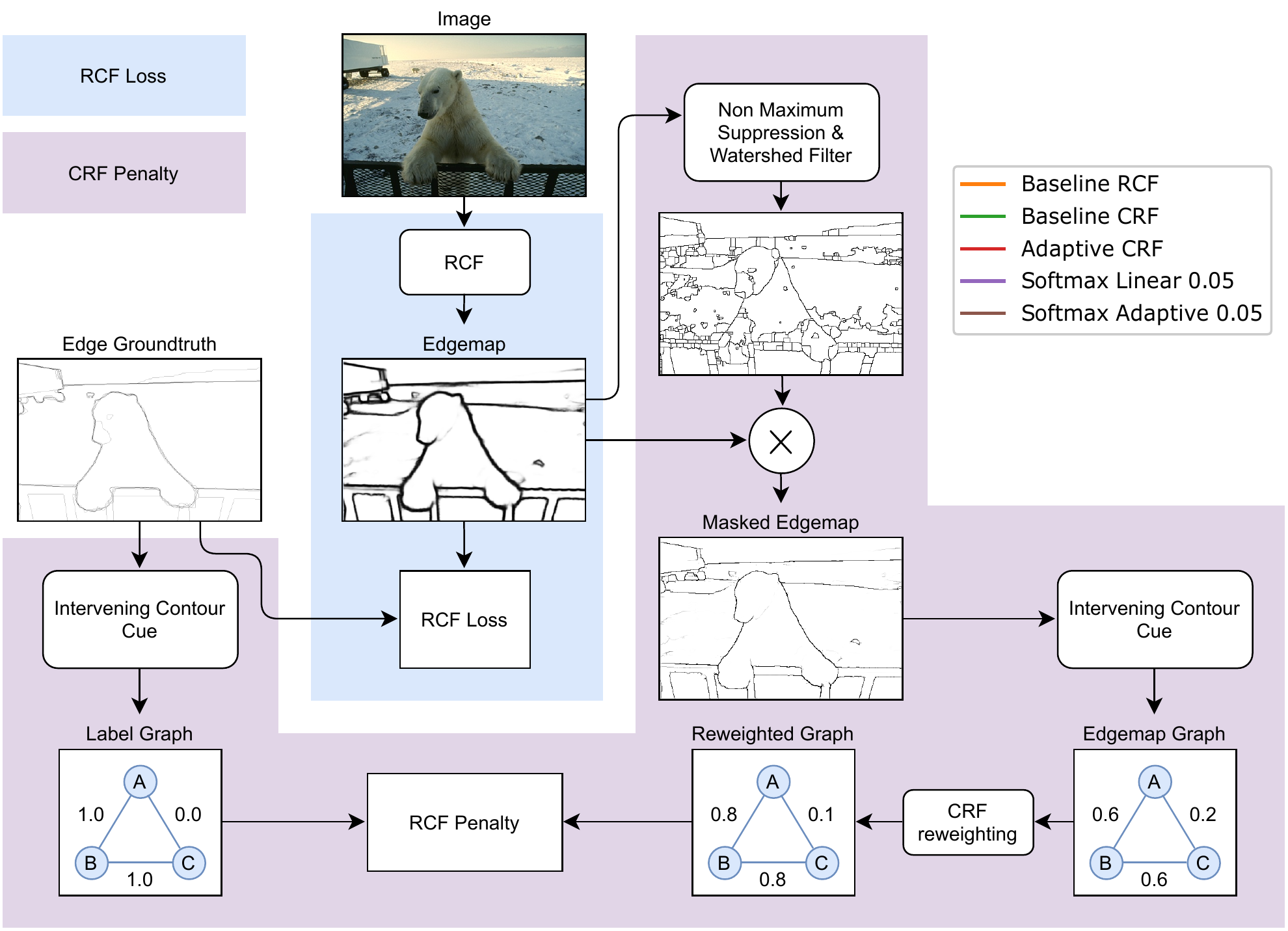}} & \\ 
          & \includegraphics[height=3cm]{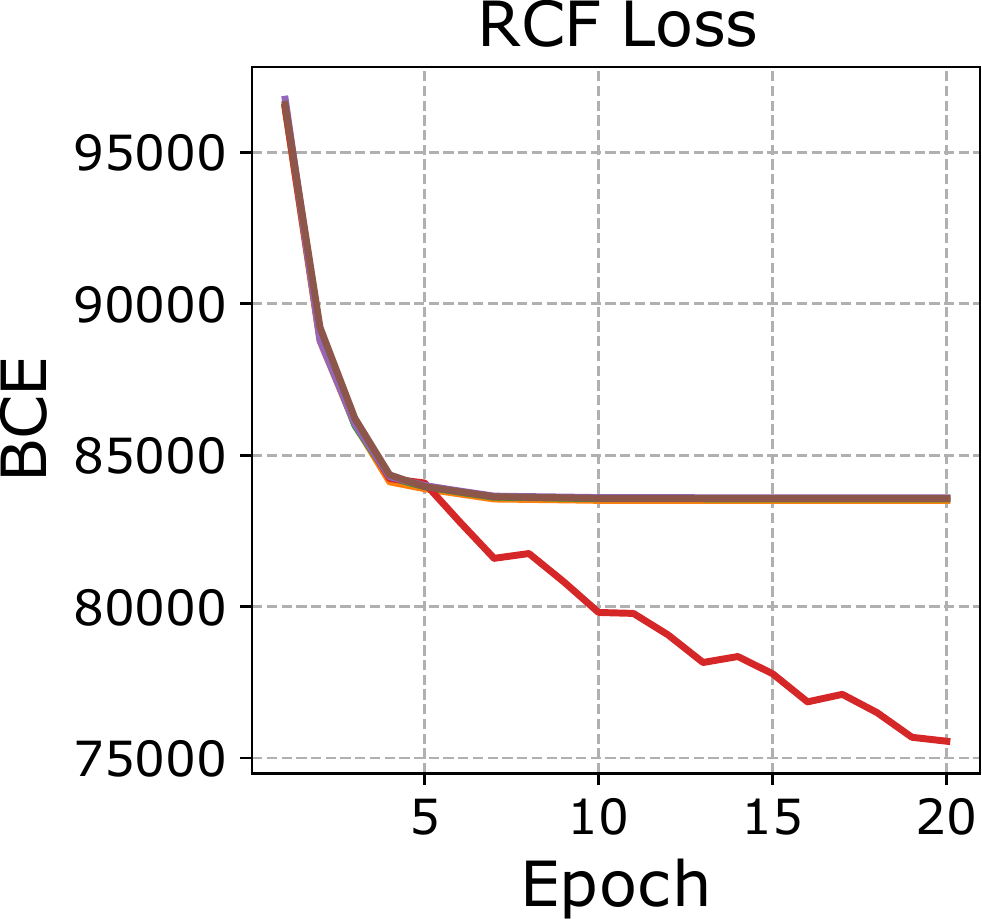} \\
          & (b) RCF Loss. \\ \\
          & \includegraphics[height=3cm]{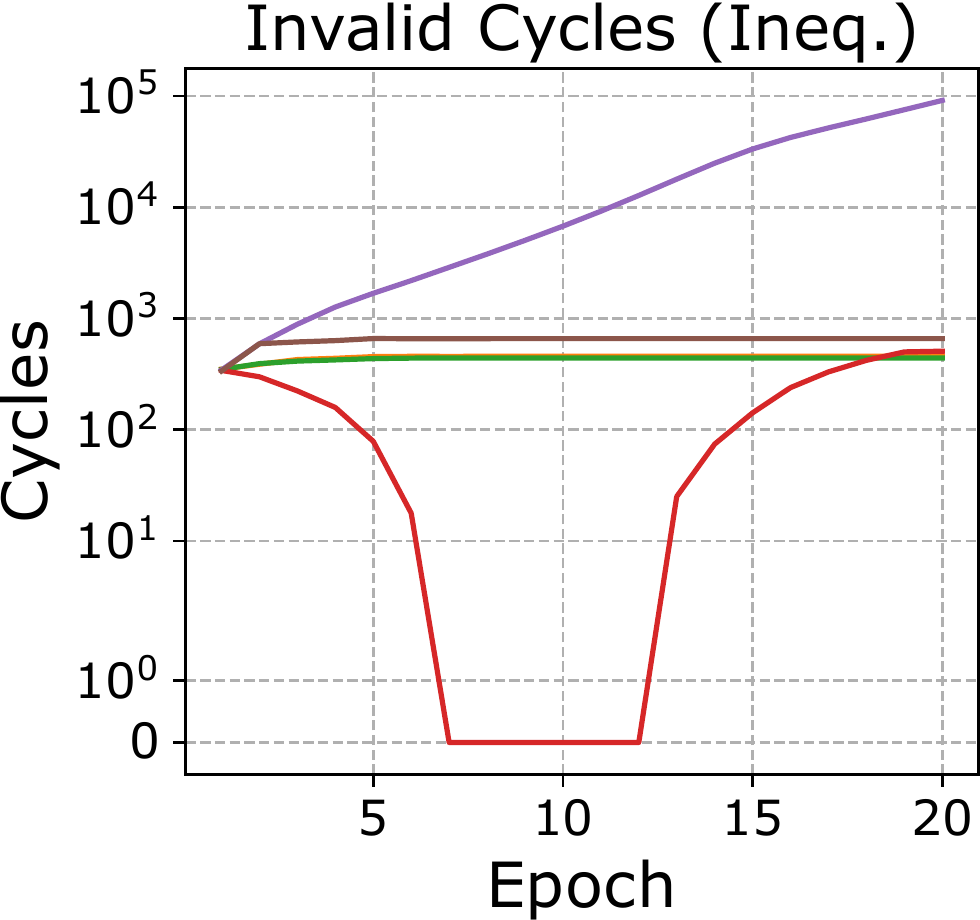} \\
          & (c) Invalid Cycles. \\ \\
          & \includegraphics[height=3cm]{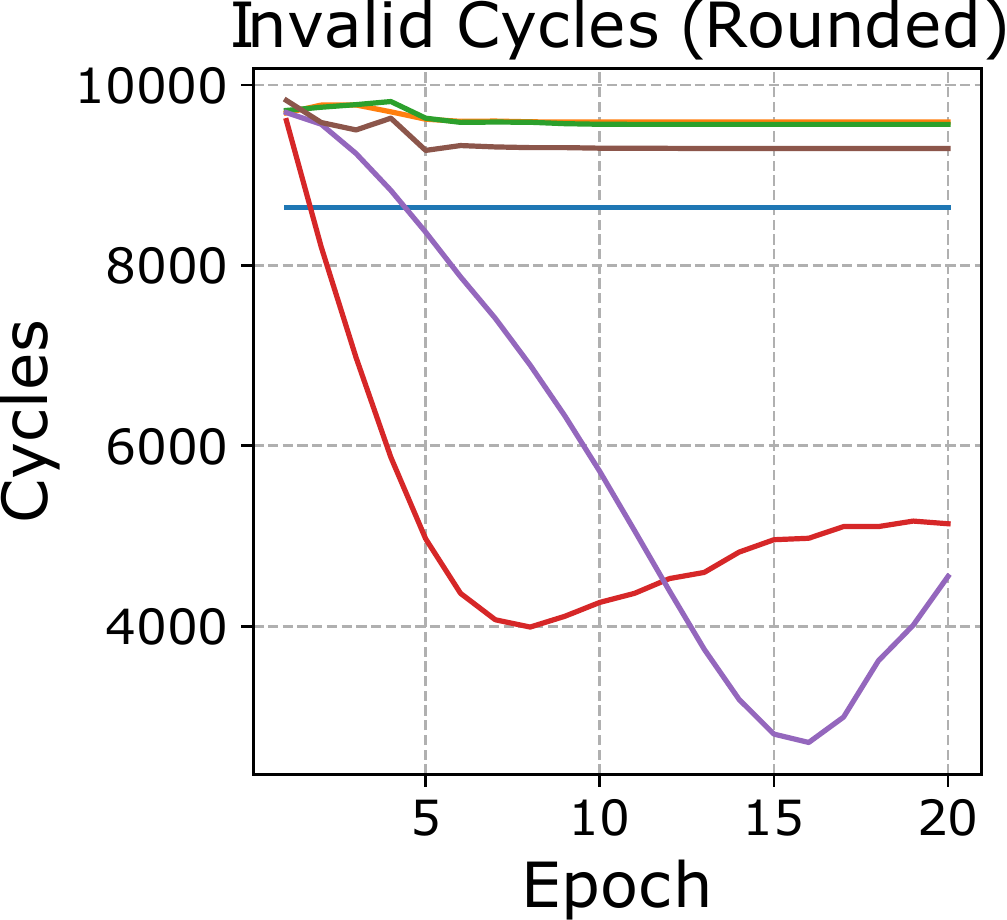} \\
          (a) The RCF-CRF training process. & (d) Invalid Cycles (rounded). \\
    \end{tabular}
    \caption{
        (a) The RCF-CRF training process is depicted, where blue highlights the RCF loss and purple highlights the additional CRF penalty.
        (b) Training progress in terms of RCF loss when training in different settings.
        Only Adaptive CRF is able to reduce the RCF loss further, while all other settings provide insufficient training signal to further improve on the basic RCF model.
        (c) Number of invalid cycle inequalities during training.
        Adaptive CRF significantly outperforms all other training settings.
        (d) Number of invalid cycle inequalities after rounding the solution during training.
        Adaptive CRF and Softmax CRF are able to improve on baselines significantly.
    }
    \label{rcf-crf}
\end{figure*}

The Richer Convolutional Features (RCF) architecture for edge detection has been recently developed by \citet{rcf-2019}.
Their main idea is based on Holistically-nested edge detection (HED) \cite{hed}, where an image classification architecture like VGG16 \cite{vgg} is divided into five stages.
Each stage produces a side output while and fully connected layers are removed.
These side outputs are trained with an individual loss function and combined by a weighted fusion layer that learns to combine them.
In contrast to HED, which only considers the last convolutional layer from each stage, the RCF architecture uses all layer information.
Hence, the features used become "richer".
Every side output is transformed via sigmoid activation, creating edge probability maps.

\subsection{RCF-CRF Architecture}
Fig. \ref{rcf-crf} shows the full proposed architecture where the CRF is combined with the RCF to optimize for consistent boundary predictions.
In a first step edge maps $Pb$ from the input image are computed by the RCF, applying the RCF Loss during training. To apply the CRF Penalty, we then need to efficiently generate a pixel-graph that represents boundaries as edge weights. For every pixel, edges of the 8-connectivity are inserted for all distances in the range of 2 to 8 pixels yielding a significant amount of cycles. For efficient weight computation, we compute Watershed boundaries on the non-maximum suppressed RCF output of the pretrained RCF model \emph{once}. We can use those to mask subsequent RCF predictions and efficiently retrieve graph weights as they are updated by the network.
As edge weights, we consider the Intervening Contour Cue (ICC) \citet{leung1998contour} which computes the probability of an edge between pixels $i$ and $j$ to be cut as  $W_{i,j}=max(Pb(x)_{x\in L_{i,j}})$,
where $x$ represents a coordinate in the image, $Pb(x)$ is the edge probability at $x$ and $L_{i,j}$ indicates the set of all coordinates on the line between $i$ and $j$ including themselves. From the Watershed masks, potential locations of the maximum can be pre-computed for efficiency.

Edge ground truth labels are created on the fly similar to the ICC by checking if there is an edge on the line between two pixels in the ground truth edge probability map to determine their cut/join label. 
The RCF loss is based on the cross entropy used in the original RCF \cite{rcf-2019}, which ignores controversial edge points that have been annotated as an edge by less than half of the annotators but at least one.
They are summed to obtain the final RCF-CRF loss.
After training, the model segmentations can be computed using hierarchical approaches such as MCG \cite{APBMM2014} or multicut solvers such as \cite{keuper_lifted}.

\section{Experiments}
We evaluate our approach in two different image segmentation applications.
First, we show experiments and results on the BSDS500 \cite{bsds} dataset for edge detection and image segmentation.
Second, we consider the segmentation of bio-medical data, in particular, electron microscopic recordings of neuronal structures of fruit flies~\cite{isbi}.
\begin{figure}[t]
	\centering
	\begin{tabular}{c@{ }c@{ }c@{ }c@{ }c}
	    \includegraphics[width=0.135\textwidth]{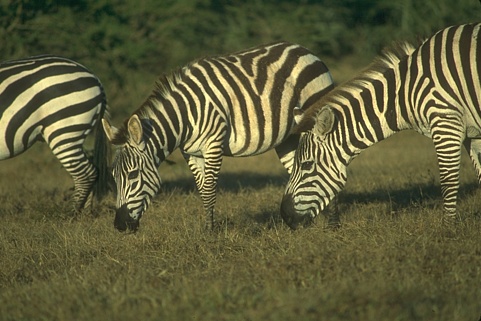}
	     &\multirow{2}{*}{\rotatebox{90}{RCF}}
	     &\includegraphics[width=0.135\textwidth]{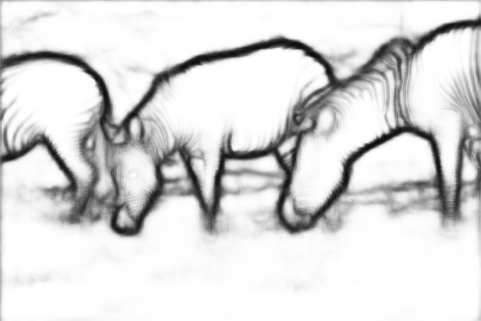} 
		&\multirow{2}{*}{\rotatebox[origin=c]{90}{ours}}
&		\includegraphics[width=0.135\textwidth]{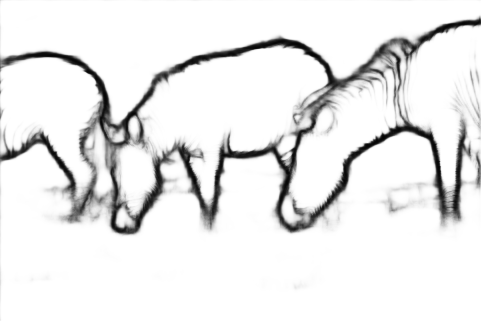} \\  
		&&\includegraphics[width=0.135\textwidth]{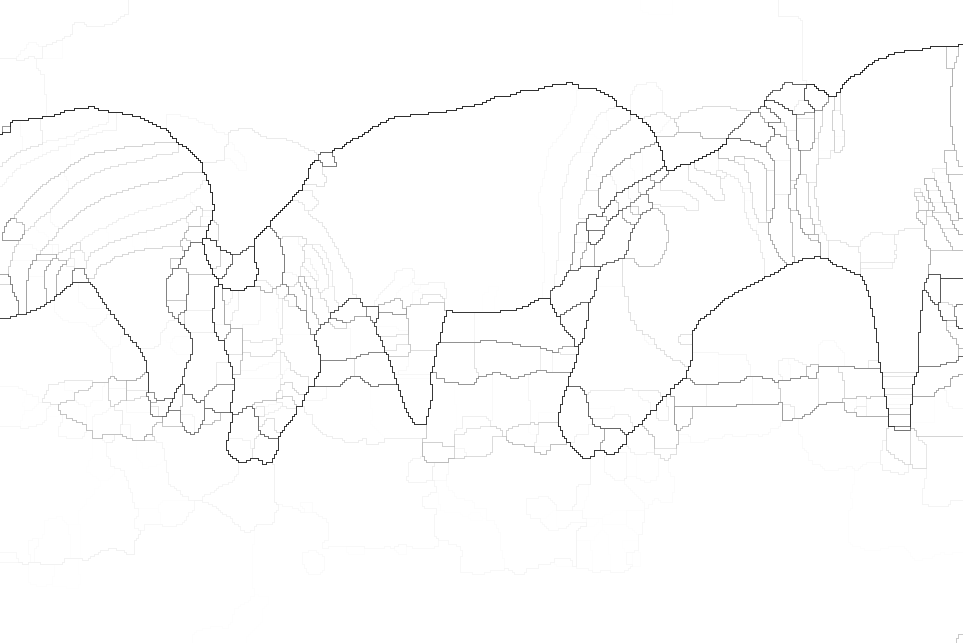} &&
		\includegraphics[width=0.135\textwidth]{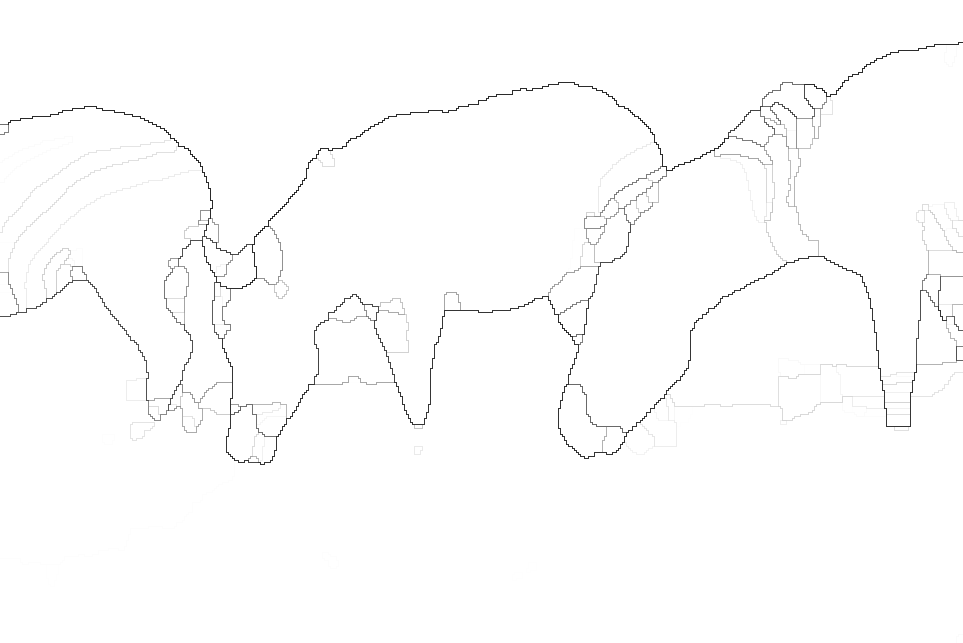} \\
	\end{tabular}
	\caption{
	Example BSDS500 \cite{bsds} test image. Top row depicts original image, Baseline RCF and Adaptive CRF edge maps.
	Bottom row depicts the respective UCMs.
	The edge map optimized with Adaptive CRF is less cluttered while accurately localizing the contours.
	}
	\label{ex-fig}
\end{figure}
\subsection{Berkeley Segmentation Dataset and Benchmark}
BSDS500~\cite{bsds} contains $200$ train, $100$ validation and $200$ test images that show colored natural photography often depicting animals, landscapes, buildings, or humans.
Due to its large variety in objects with different surfaces and different lightning conditions, it is generally considered a difficult task for edge detection as well as image segmentation.
Several human annotations are given per image.
The RCF is pre-trained on the augmented BSDS500 data used by \citet{rcf-2019} and \citet{hed}.
After convergence it is fine-tuned with our CRF using only the non-augmented BSDS images to reduce training time.
We evaluate the impact of fine-tuning the CRF in different training settings.
These are: (Baseline RCF) further training of the RCF without CRF, (Baseline CRF) fine-tuning the RCF network without cooling scheme, and (Adaptive CRF) fine-tuning with cooling scheme.
Additionally, we consider two settings where we enforce more binary solutions by introducing temperature decay in the softmax \cite{hinton2015distilling} that is applied after each mean field iteration.
Here, we consider two cases: (Softmax Linear) where we decay the temperature by $0.05$ after each epoch, and (Softmax Adaptive), where we consider the same adaption process as in in Eq.~\eqref{eq:cool}.

\paragraph{Invalid Cycles}

Fig. \ref{rcf-crf}(c) depicts the number of violated cycle inequalities during fine-tuning of the RCF network in different settings.
There, it can be seen that Baseline CRF is able to reduce the number of violations rapidly to around $500$ invalid cycles per image on average.
During training, it constantly stays at this level and does not significantly change anymore.
Due to this observation we set parameter $a$ in the cooling scheme of Adaptive CRF and softmax temperature decay to $500$. 
When training Adaptive CRF with this parameter setting, the number of violations further decreases to zero.
The impact of reduced uncertainty can furthermore be seen when looking at the number of invalid cycles after rounding edge probabilities to binary edge labels.
As Fig. \ref{rcf-crf}(d) shows, only the settings Adaptive CRF and Softmax Linear are able to reduce the number of invalid cycles after rounding.
However, Softmax Linear is not able to reduce the number of violations before rounding in contrast to Adaptive CRF.
This indicates that the cooling scheme of Adaptive CRF provides a better training signal for the RCF, which is also confirmed considering RCF training loss depicted in Fig. \ref{rcf-crf}(b).
Only Adaptive CRF is able to provide sufficiently strong training signals such that the RCF network can further decrease its training loss.
Interestingly, the number of invalid rounded cycles as well as constraint violations start increasing again after some training iterations for the adaptive CRF.
This shows the trade-off between the RCF loss and the penalization provided by the CRF.

\paragraph{Edge Detection}

Tab. \ref{bsds-edge-table} shows evaluation scores on the BSDS500 test set.
The F-measure at the optimal dataset scale (ODS) and the optimal image scale (OIS) as well as the Average Precision (AP) are reported.
The multiscale version (MS) is computed similar to \cite{rcf-2019} with scales $0.5$, $1.0$ and $1.5$.
Adaptive CRF achieves significant improvements compared in terms of ODS and OIS to all other training settings. The respective precision recall curves are given in the Appendix. AP decreases slightly with the CRF models, which is expected since the CRF removes uncertain edges that do not form closed components and therefore affects the high recall regime.
A qualitative example is shown in Fig. \ref{ex-fig} and in the Appendix.
\begin{table}[t!]
	\centering
	\caption[BSDS Edge Detection Results]{
	Edge Detection Results on the BSDS500 test set.
	All RCFs are based on VGG16.
	Results reported for Baseline RCF are computed for a model trained by ourselves and are slightly worse than the originally reported scores in \cite{rcf-2019}.
	}
	\label{bsds-edge-table}
	\begin{tabular}{lccc}
		\toprule
		\textbf{Models} & \textbf{ODS} & \textbf{OIS} & \textbf{AP}  \\
		\midrule
        Baseline RCF                    &         $0.811$  &         $0.827$  &         $0.815$  \\
		Baseline RCF (MS)               &         $0.812$  &         $0.830$  & $\mathbf{0.836}$ \\
		Baseline CRF (ours)             &         $0.810$  &         $0.827$  &         $0.815$  \\
		Baseline CRF (MS) (ours)        &         $0.812$  &         $0.831$  & $\mathbf{0.836}$ \\
		Softmax Linear (ours)           &         $0.810$  &         $0.826$  &         $0.815$  \\
		Softmax Linear (MS) (ours)      &         $0.812$  &         $0.831$  & $\mathbf{0.836}$ \\
		Adaptive CRF (ours)             &         $0.815$  &         $0.830$  &         $0.812$  \\
		Adaptive CRF (MS) (ours)        & $\mathbf{0.817}$ & $\mathbf{0.835}$ &         $0.833$  \\
		\bottomrule
	\end{tabular}
\end{table}

\begin{table}[t!]
	\centering
	\caption[BSDS Segmentation Results]{
	Segmentation results on the BSDS500 test set.
	All RCFs are based on VGG16.
	Results reported for Baseline RCF are computed for a model trained by ourselves and are slightly better than the scores reported in \cite{rcf-2019}.
	}
	\label{bsds-seg-table}
	\begin{tabular}{lccc}
		\toprule
		\textbf{Models} & \textbf{ODS} & \textbf{OIS} & \textbf{AP}  \\
		\midrule
        Baseline RCF                    &         $0.803$  &         $0.829$  &         $0.832$  \\
		Baseline RCF (MS)               &         $0.808$  &         $0.832$  & $\mathbf{0.849}$ \\
		Baseline CRF (ours)             &         $0.804$  &         $0.828$  &         $0.831$  \\
		Baseline CRF (MS) (ours)        &         $0.808$  &         $0.831$  & $\mathbf{0.849}$ \\
		Softmax Linear (ours)           &         $0.803$  &         $0.829$  &         $0.831$  \\
		Softmax Linear (MS) (ours)      &         $0.808$  &         $0.831$  & $\mathbf{0.849}$ \\
		Adaptive CRF (ours)             &         $0.808$  &         $0.830$  &         $0.828$  \\
		Adaptive CRF (MS) (ours)        & $\mathbf{0.813}$ & $\mathbf{0.834}$ &         $0.847$  \\
		\bottomrule
	\end{tabular}
\end{table}

\paragraph{Image Segmentation}

To obtain a hierarchical segmentation using the predicted edge maps we compute MCG \cite{APBMM2014} based Ultrametric Contour Maps (UCM) \cite{bsds} that generate hierarchical segmentations based on different edge probability thresholds. Edge orientations needed for MCG were computed using the standard filter operations. 
In contrast to \cite{rcf-2019} we do not use the COB framework but use pure MCG segmentations to allow for a more direct assessment of the proposed approach.
Results for all training settings are reported in Tab.~\ref{bsds-seg-table}.
Again, the Adaptive CRF models outperforms all other models in ODS, while AP is only slightly affected. Multi-scale (MS) information additionally improves results. 

\begin{figure}[t!]
	\centering
	\includegraphics[width=0.45\textwidth]{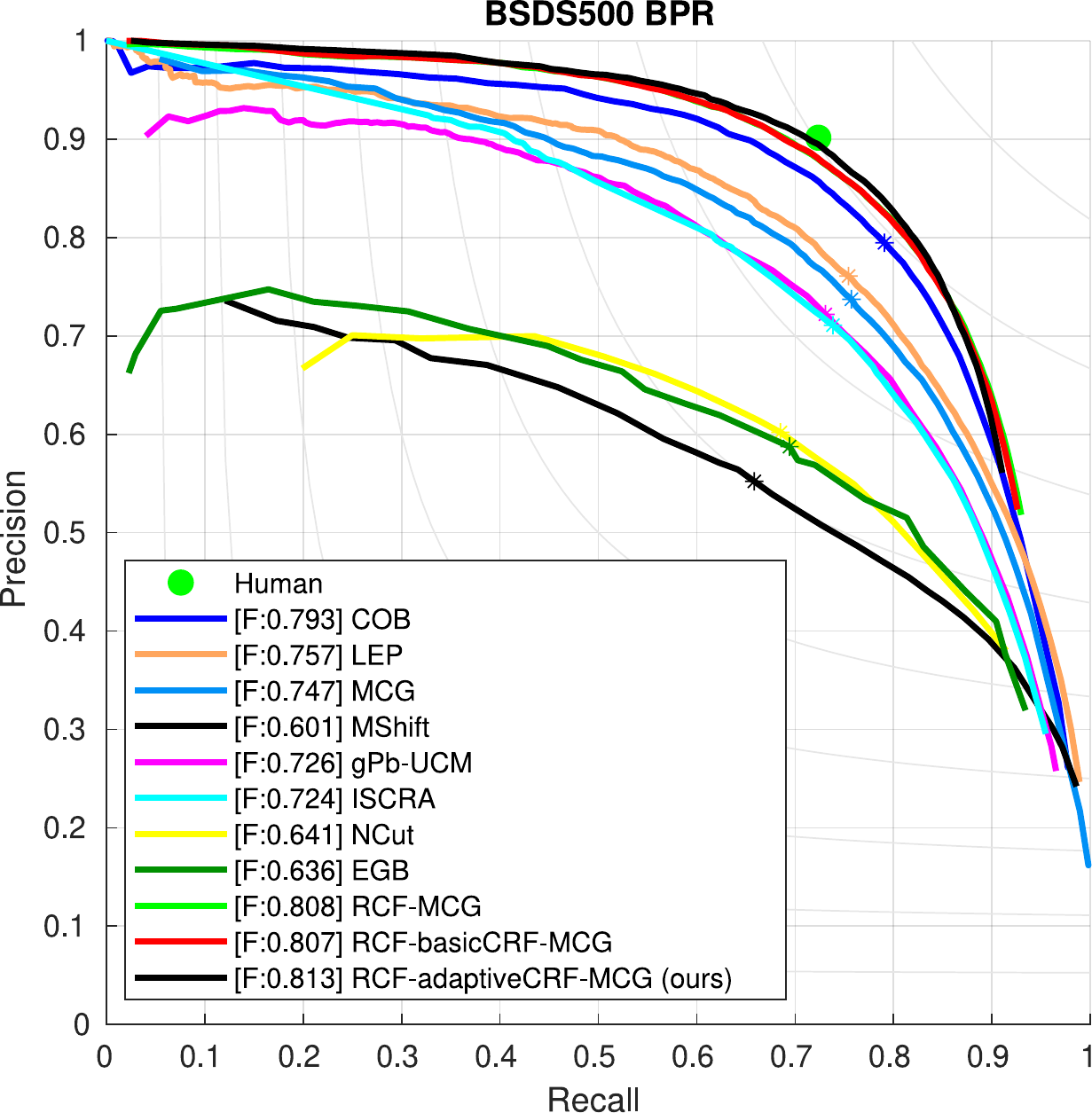}
	\caption[BSDS Segmentation PR Curves]{Precision recall curves for segmentation on the BSDS500 test set. The proposed RCF with Adaptive CRF yields the highest ODS score and operates at a higher precision level than other models.}
	\label{bsds-seg-vgg-pr}
\end{figure}

Fig. \ref{bsds-seg-vgg-pr} depicts the segmentation PR curves comparing the RCF based methods to other standard models.
Similar to the edge map evaluation (see Appendix) the curves are steeper for the CRF based model compared with the plain RCF, thereby following the bias of human annotations, i.e., approaching the green marker in Fig. \ref{bsds-seg-vgg-pr}.
Depending on when the curves start to tilt the corresponding F-measure can be slightly lower as it is the case for the basic CRF.
Adaptive CRF, however, also yields a considerably higher F-score improving over the baseline from $0.808$ to $0.813$.
This result shows that
employing cycle information is generally beneficial to estimate closed boundaries.

\subsection{Neuronal Structure Segmentation}

Next, we conduct experiments on the segmentation of neuronal structures~\cite{isbi}. The data was obtained from a serial section Transmission Electron Microscopy dataset of the Drosophila first instar larva ventral nerve cord \cite{cardona2010integrated, cardona2012trakem2}. This technique captures images of the Drosophila brain with a volume of 2 $\times$ 2 $\times$ 1.5$\mu$ and a resolution of 4 $\times$ 4 $\times$ 50 nm/pixel. The volumes are anisotropic, i.e. while the x- and y-directions have a high resolution the z-direction has a rather low resolution \cite{isbi}. 
Both, training and test set consist of a stack of 30 consecutive grayscale images. The goal of the challenge is to produce a binary map that corresponds to the membranes of cells in the image.

\citet{thorsten} have applied a multicut approach to this application. Their pipeline first produced an edge map from the original images by using either a cascaded random forest or a CNN.
In order to reduce the complexity they aggregated individual pixels to superpixels by using the distance transform watershed technique \cite{acharjya2013new}. Based on these superpixels they solve the multicut and the lifted multicut problem using the fusion moves algorithm \cite{beier2015fusion}.
\begin{table}[t!]
	\centering
	\caption[ISBI Testset Results]{Results of the ISBI challenge on the test set.}
	\label{isbi-res}
	\begin{tabular}{lcc}
		\toprule
		\textbf{Models}
		& $ \mathbf{V^{\mathrm{Rand}}} $ & $ \mathbf{V^{\mathrm{Info}}} $  \\
		\midrule
		
			Baseline \cite{thorsten} &0.9753 &0.9874\\
		Basic-CRF-optimized (ours)&0.9784 &\bf{0.9889} \\		Adaptive-CRF-optimized (ours)&\bf{0.9808} & 0.9887\\

		\bottomrule
	\end{tabular}
\end{table}

In this context, we apply Adaptive CRF as a post-processing method to an existing graph without training the underlying edge detection.
Edge weights for the test set are computed using a random forest in the simple (not lifted) multicut pipeline from \cite{thorsten} and define a graph.
We optimize these graph weights with the proposed approach and subsequently decompose the graph using the fusion move algorithm as in \cite{thorsten}.
The number of mean-field iterations was set to $20$ and for the adaptive CRF the update threshold $a$ was set to $100$.
Since the CRF is not trained but used only for optimizing the graph once, the update threshold is evaluated after every mean-field iteration rather than every epoch.

The graph obtained from \citet{thorsten} for the test set contained $74\,485$ cycles in total.
Before applying the CRF, $61\,099$ of them violated the cycle inequality constraints and $38\,283$ cycles were invalid after rounding.
Afterwards, both cycle counts were close to zero.
Tab. \ref{isbi-res} contains the results obtained for the not-modified edge weights (Baseline), the edge weights optimized with Baseline CRF and the edge weights optimized with Adaptive CRF.
For evaluation we also refer to the ISBI challenge \cite{isbi} that indicates two measures: the foreground-restricted Rand Scoring after border thinning $ V^{\mathrm{Rand}} $ and the foreground-restricted Information Theoretic Scoring after border thinning $ V^{\mathrm{Info}} $.
Both CRF models were able to improve the segmentation result in both evaluation metrics.
While the differences in $ V^{\mathrm{Info}} $-score are rather small, Adaptive CRF increased the $ V^{\mathrm{Rand}} $-score by $0.005$ compared to the baseline.
Taking into account that the baseline is already very close to human performance,
this is a very good result.
Comparing the two CRF models it can be seen that the $ V^{\mathrm{Info}} $-score is almost the same for both approaches.
In terms of $ V^{\mathrm{Rand}} $-score, Adaptive CRF improves stronger over the baseline than Baseline CRF.
Overall, this experiment shows that applying the CRF is beneficial for image segmentation even without training the edge extraction. Accordingly, our approach can be applied even as a post-processing step without increasing training time.

\section{Conclusion}

We introduce an adaptive higher-order CRF that can be optimized jointly with an edge detection network and encourages edge maps that comply with the cycle constraints from the Minimum Cost Multicut Problem.
Combining the CRF with the RCF model \cite{rcf-2019} for edge detection yields much sharper edge maps and promotes closed contours on BSDS500. PR curves show that the CRF based model yields steeper curves having a higher precision level.
Similarly, the resulting segmentations show that the approach is able to generate more accurate and valid solutions.
Moreover, the CRF can be used as post-processing to optimize a graph for cycle constraints as shown on the electron microscopy data.
It has shown considerable improvement in the evaluation metrics without increasing training time.

\bibliography{aaai22}

\newpage
\onecolumn
\section*{Optimizing Edge Detection for Image Segmentation with Multicut Penalties:\\Supplementary Material}

\subsection*{Appendix A: Training Details}

\paragraph{Adaptive CRF}

\begin{figure}[H]
	\centering
	\begin{tabular}{ccc}
	    \includegraphics[width=0.35\textwidth]{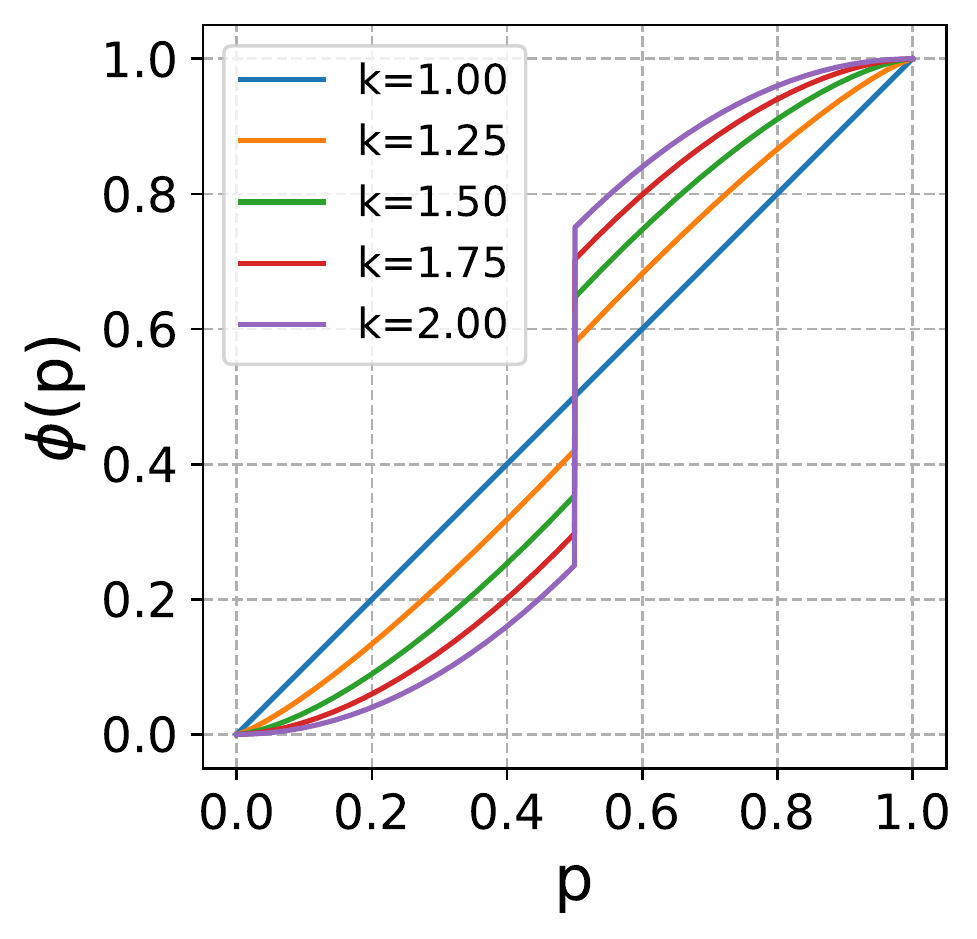} &
        \includegraphics[width=0.35\textwidth]{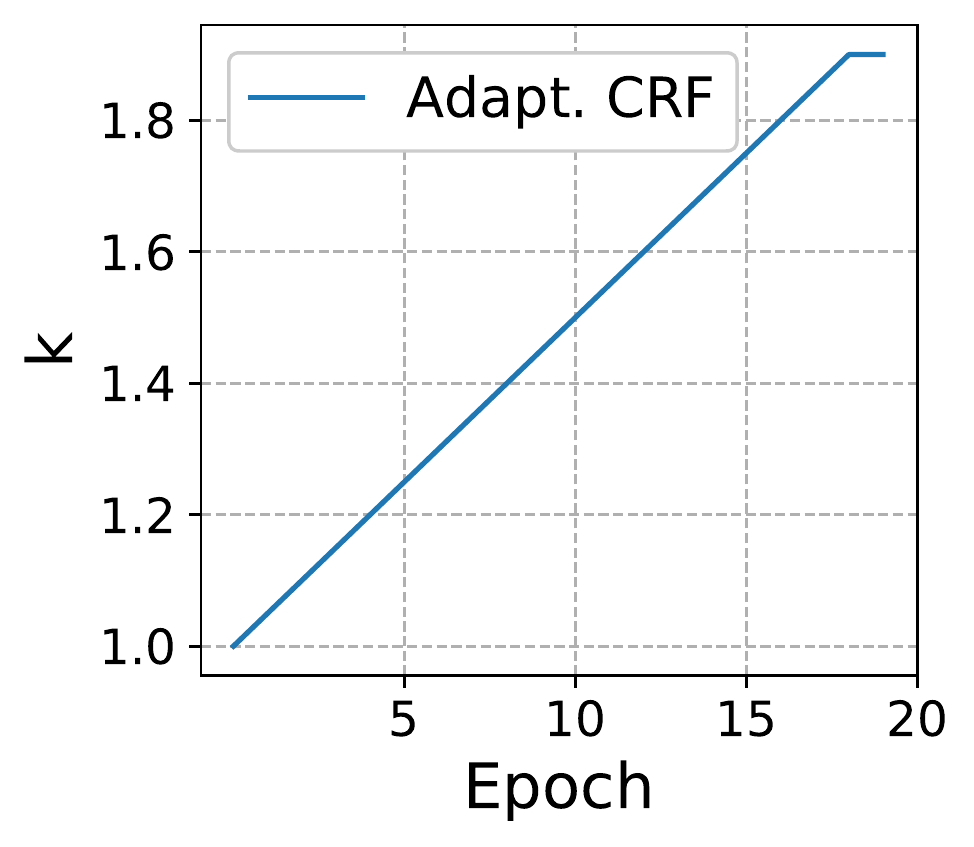}
	\end{tabular}
	\caption{
        (left) Function $\phi$ plotted for different values of $k$.
        (right) Value of $k$ during $20$ epochs of training Adaptive CRF.
	}
	\label{fig:supp-adaptive-crf}
\end{figure}

By reformulating mean field updates from Eq. \eqref{mean-field-update} with our adaptive function $\phi$ (see Eq. \eqref{eq:phi}), we get:

\begin{multline}
\label{new_mean-field-update}
Q^{t}_{i}(y_{i}=l)= 
\frac{1}{Z_{i}}exp \bigggl \{
-\sum \limits_{c\in C} \bigggl ( \\
\sum \limits_{p\in P_{c|y_{i}=l}} \biggl (\prod \limits_{j\in c,j\neq i}\phi\left(Q^{t-1}_{j}(y_{j}=p_{j}),k\right)\biggr )
\gamma_{p}
\,\,+ 
\\ \biggl( 1- \biggl( \sum \limits_{p\in P_{c|x_{i}=l}}\biggl (\prod \limits_{j\in c,j\neq i}\phi\left(Q^{t-1}_{j}(y_{j}=p_{j}),k\right)\biggr )
	\biggr )
	\biggr )\gamma_{max}
\\ \bigggr )
\bigggr \},
\end{multline}
where function $\phi$ modifies probabilities such that their values are pushed closer to $0$ or $1$, depending on their current value.
Fig.~\ref{fig:supp-adaptive-crf}(a) shows a plot of this function with different values of $k$.
During training, $k$ is initialized with $1.0$ and then increased based on the cooling scheme defined in Eq.~\eqref{eq:cool}.
When training Adaptive CRF on BSDS500 for $20$ epochs, we observe that $k$ is increased each epoch with the exception of the last one (see Fig.~\ref{fig:supp-adaptive-crf}(b)).

\begin{figure}[H]
	\centering
	\begin{tabular}{c}
        \includegraphics[width=0.33\textwidth]{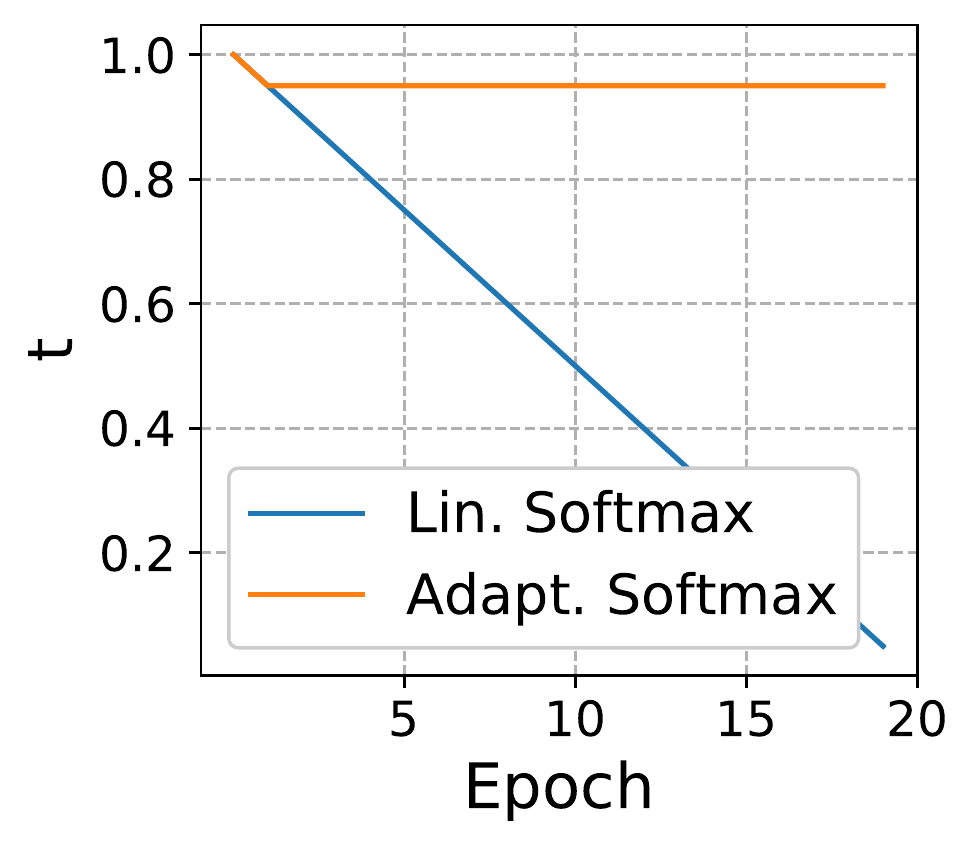}
	\end{tabular}
	\caption{
	Values of $t$ during $20$ epochs of training Linear Softmax and Adaptive Softmax.
	}
	\label{fig:softmax-t}
\end{figure}

\paragraph{Linear Softmax and Adaptive Softmax}
The purpose of $\phi$ is to push values towards $1$ or $0$, and therefore enforce multicut constraints more strictly.
However, this should also be possible by considering temperature decay in the softmax function that is called after each mean field update iteration.
Hence, instead of
\begin{equation}
    Q^t_i(y_i=l) = \frac{\exp\bigl\{\tilde{Q}^t_i(y_i=l)\bigr\}}
                        {\sum_l \exp\bigl\{\tilde{Q}^t_i(y_i=l)\bigr\}}
\end{equation}
we compute
\begin{equation}
    Q^t_i(y_i=l) = \frac{\exp\bigl\{\tilde{Q}^t_i(y_i=l) / t \bigr \}}
                        {\sum_l \exp\bigl\{\tilde{Q}^t_i(y_i=l) / t \bigr\}},
\end{equation}
where we initialize $t=1.0$ and decay each epoch in the case of Linear Softmax by $0.05$.
In the case of Adaptive Softmax we use the same condition as for Adaptive CRF (see Eq.~\eqref{eq:cool}), where $t$ is only decayed by $0.05$ if the average number of violated cycle inequalities is $\leq 500$.
Fig.~\ref{fig:softmax-t} shows the temperature decaying schemes during training of $20$ epochs on BSDS500.

\paragraph{Training Runs}
We train each model on one RTX8000 GPU with batch size $10$, and optimize with SGD with learning rate $10^{-6}$, momentum $0.9$, weight decay $2\cdot10^{-4}$, stepsize $3$, gamma $0.1$ for $20$ epochs.
Code is provided with the supplementary submission.

\subsection*{Appendix B: Quantitative Edge Detection Results on BSDS500}

\begin{figure}[H]
	\centering
	\includegraphics[width=0.45\textwidth]{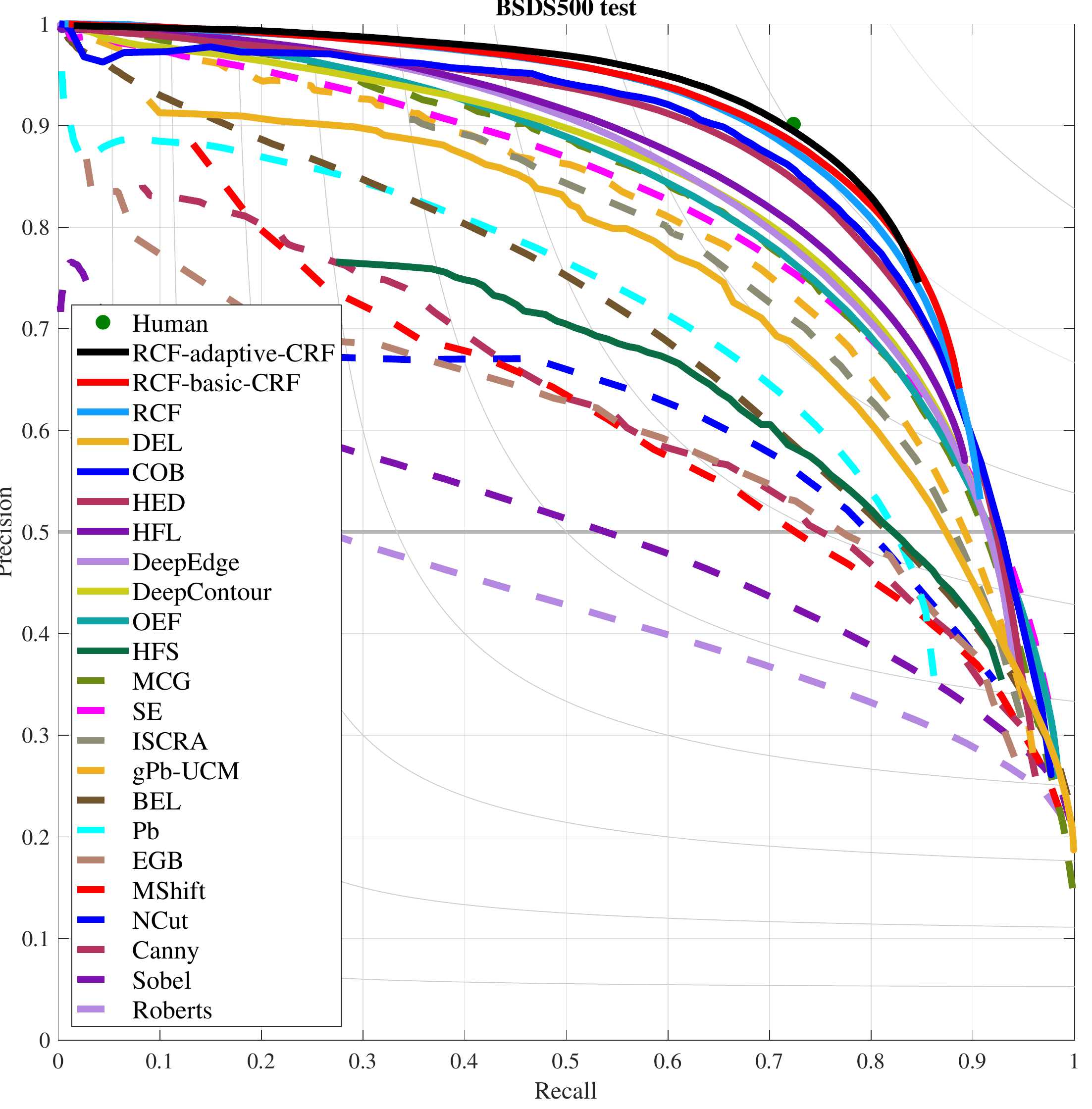}
	\caption[BSDS Edge Detection PR Curves]{Edge Detection precision recall curves for the BSDS500 test set. RCF models all have the VGG backbone. Models that were optimized with the CRF yield steeper curves.}
	\label{bsds-edge-vgg-pr}
\end{figure}

Fig.~\ref{bsds-edge-vgg-pr} shows the precision recall curves for edge detection on the BSDS500 test set.
The proposed Adaptive CRF yields the highest ODS score and operates at a higher precision level than other models.

\subsection*{Appendix C: Qualitative Results on BSDS500}
In the supplementary material we provide additional examples of BSDS500 images, their respective edge maps as well as the corresponding UCM segmentation.
We compare plain RCF results with models optimized with the basic CRF and the adaptive CRF which both encourage valid cycles that comply with the cycle inequality constraints of the Minimum Cost Multicut Problem.
It can be seen that applying the CRFs during training results in cleaner edge maps where trailing edges are removed and contours are more likely to be closed.
The effect is amplified for the adaptive CRF. 
The corresponding segmentations suffer less from oversegmented backgrounds and individual components are more precise compared to the plain RCF. 
The additional examples illustrate that promoting closed contours when learning edge maps yields more accurate segmentations.

\begin{figure}[h!]
	\centering
	
	\begin{tabular}{c@{ }c@{ }c}
	    \multicolumn{3}{c}{\includegraphics[width=0.33\textwidth]{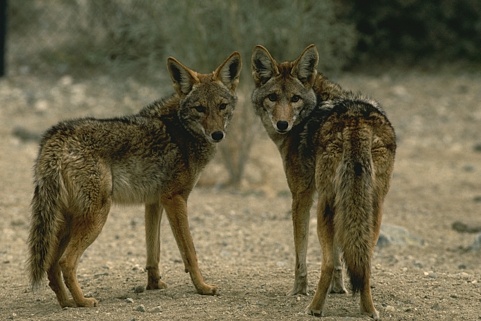}}\\
	    
	    \hline
	    \multicolumn{3}{c}{Edge maps}\\
	    \hline
	    
	    \includegraphics[width=0.33\textwidth]{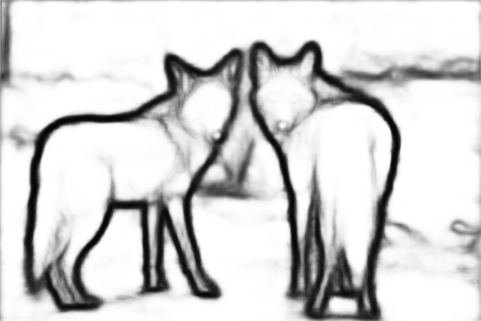}
	     &\includegraphics[width=0.33\textwidth]{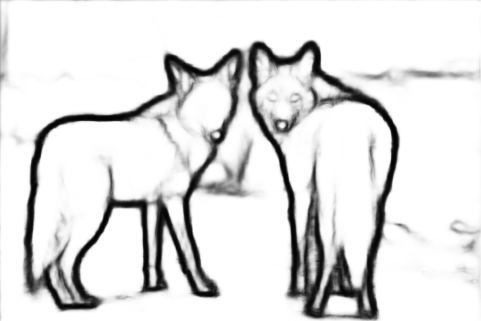}
		
&		\includegraphics[width=0.33\textwidth]{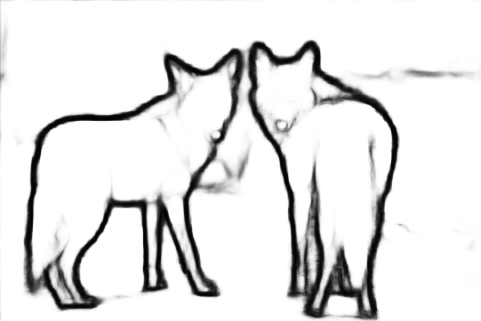} \\
\hline
	    \multicolumn{3}{c}{UCM Segmentations (Threshold=0.5)}\\
	    \hline
		\includegraphics[width=0.33\textwidth]{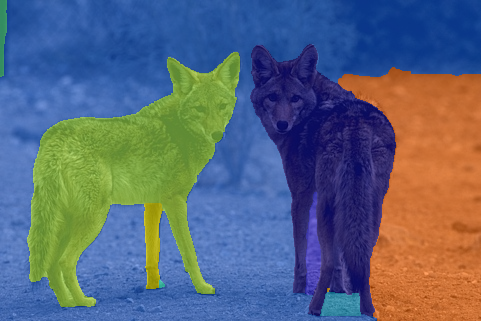}
	     &\includegraphics[width=0.33\textwidth]{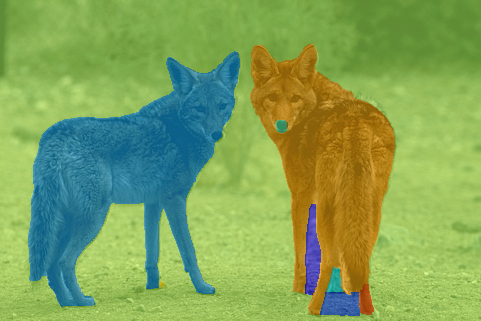}
		
&		\includegraphics[width=0.33\textwidth]{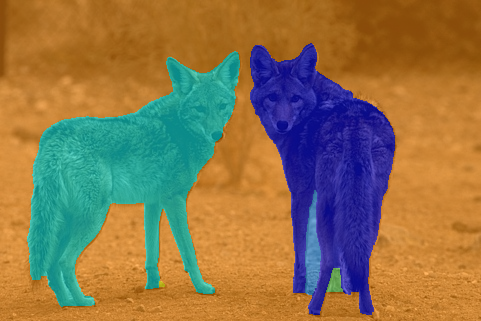} \\         \hline
	    Baseline RCF & Baseline CRF & Adaptive CRF\\
	    \hline
		
	\end{tabular}
	\label{example-fig}
\end{figure}

\begin{figure}[h!]
	\centering
	
	\begin{tabular}{c@{ }c@{ }c}
	    \multicolumn{3}{c}{\includegraphics[width=0.33\textwidth]{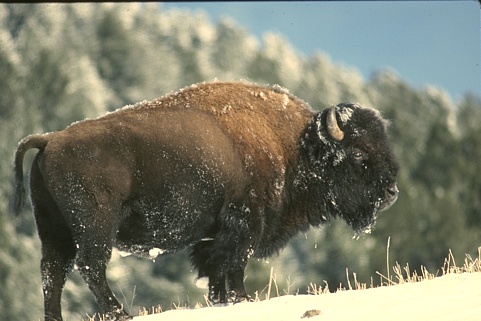}}\\
	    
	    \hline
	    \multicolumn{3}{c}{Edge maps}\\
	    \hline
	    
	    \includegraphics[width=0.33\textwidth]{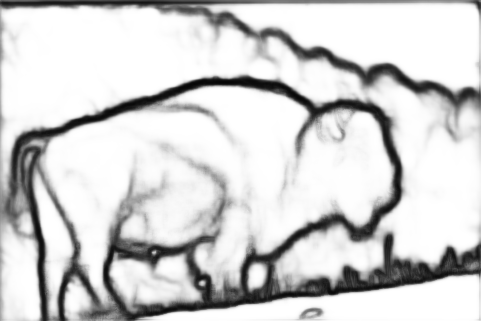}
	     &\includegraphics[width=0.33\textwidth]{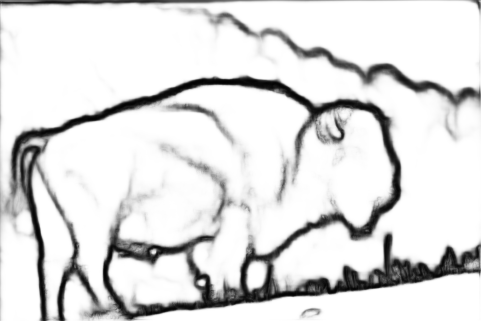}
		
&		\includegraphics[width=0.33\textwidth]{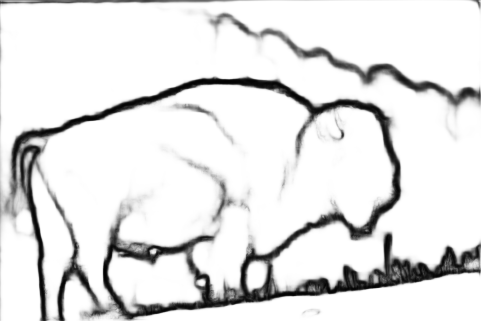} \\
\hline
	    \multicolumn{3}{c}{UCM Segmentations (Threshold=0.5)}\\
	    \hline
		\includegraphics[width=0.33\textwidth]{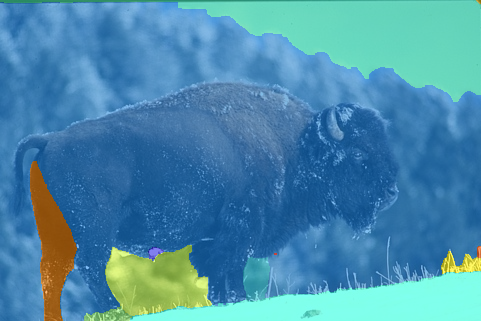}
	     &\includegraphics[width=0.33\textwidth]{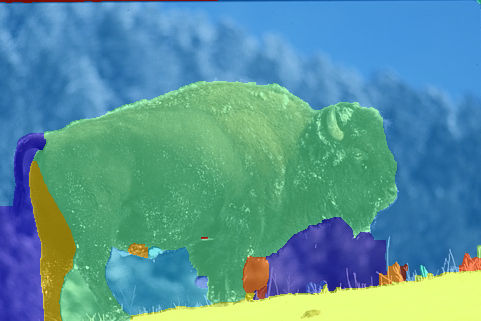}
		
&		\includegraphics[width=0.33\textwidth]{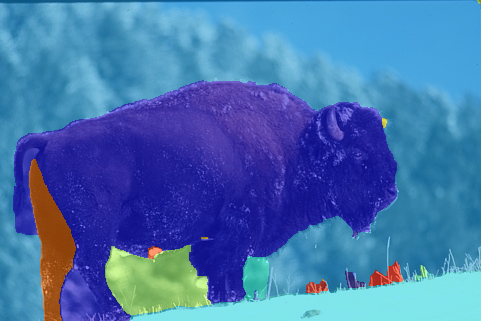} \\         \hline
	    Baseline RCF & Baseline CRF & Adaptive CRF\\
	    \hline
		
	\end{tabular}
	\label{example-fig}
\end{figure}

\begin{figure}[h!]
	\centering
	
	\begin{tabular}{c@{ }c@{ }c}
	    \multicolumn{3}{c}{\includegraphics[width=0.33\textwidth]{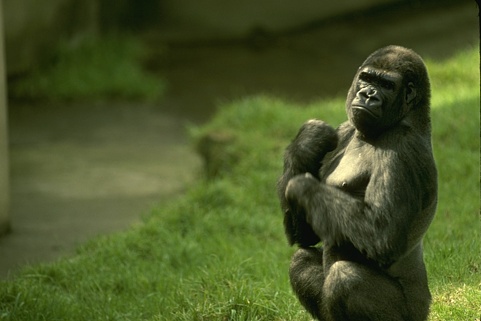}}\\
	    
	    \hline
	    \multicolumn{3}{c}{Edge maps}\\
	    \hline
	    
	    \includegraphics[width=0.33\textwidth]{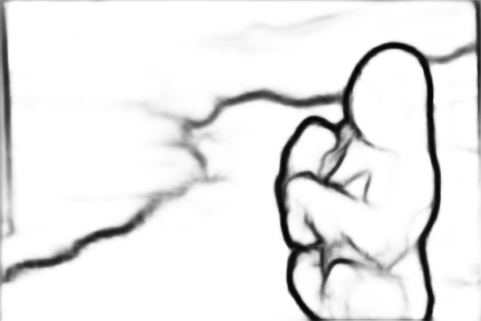}
	     &\includegraphics[width=0.33\textwidth]{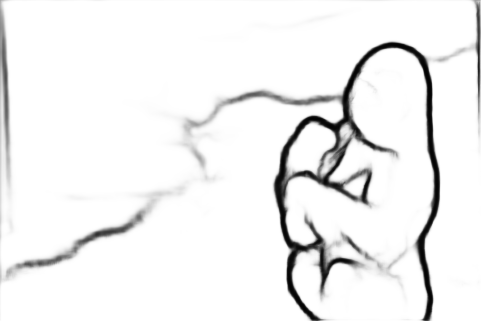}
		
&		\includegraphics[width=0.33\textwidth]{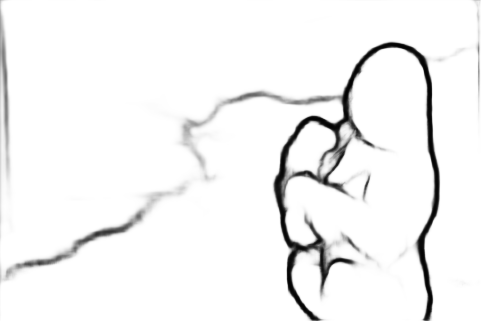} \\
\hline
	    \multicolumn{3}{c}{UCM Segmentations (Threshold=0.5)}\\
	    \hline
		\includegraphics[width=0.33\textwidth]{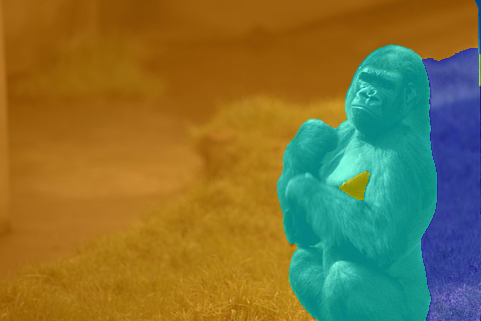}
	     &\includegraphics[width=0.33\textwidth]{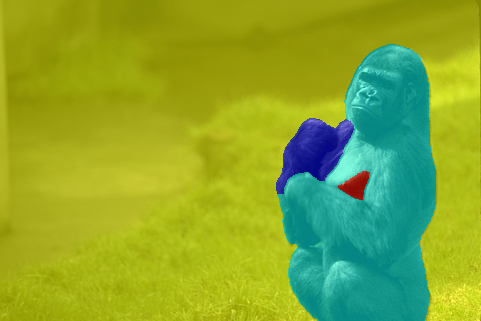}
		
&		\includegraphics[width=0.33\textwidth]{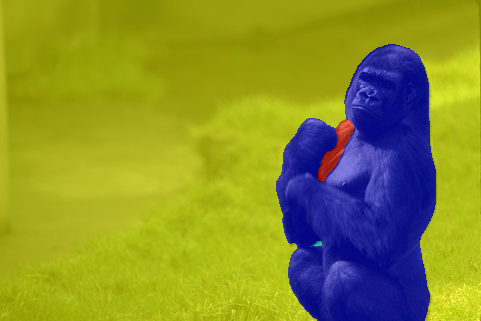} \\         \hline
	    Baseline RCF & Baseline CRF & Adaptive CRF\\
	    \hline
		
	\end{tabular}
	\label{example-fig}
\end{figure}

\begin{figure}[h!]
	\centering
	
	\begin{tabular}{c@{ }c@{ }c}
	    \multicolumn{3}{c}{\includegraphics[width=0.33\textwidth]{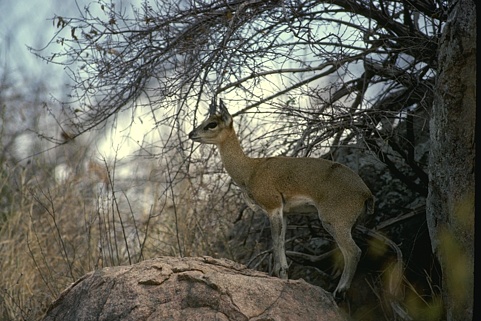}}\\
	    
	    \hline
	    \multicolumn{3}{c}{Edge maps}\\
	    \hline
	    
	    \includegraphics[width=0.33\textwidth]{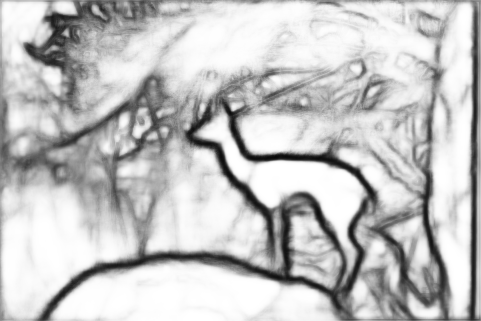}
	     &\includegraphics[width=0.33\textwidth]{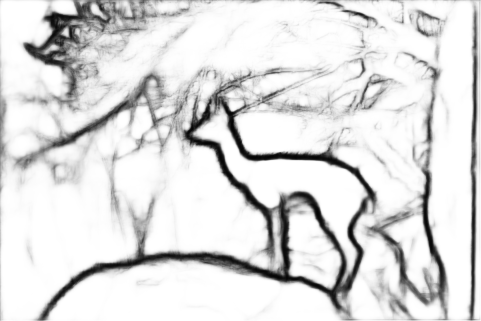}
		
&		\includegraphics[width=0.33\textwidth]{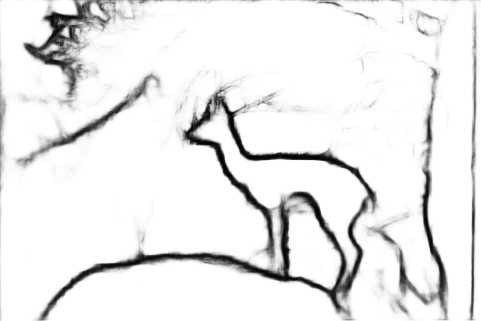} \\
\hline
	    \multicolumn{3}{c}{UCM Segmentations (Threshold=0.5)}\\
	    \hline
		\includegraphics[width=0.33\textwidth]{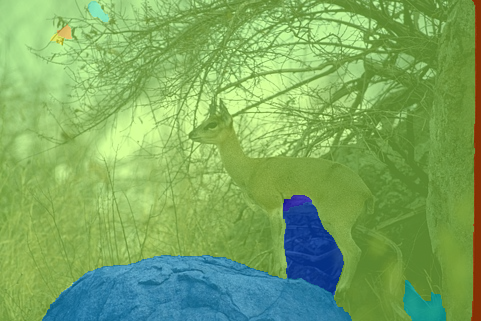}
	     &\includegraphics[width=0.33\textwidth]{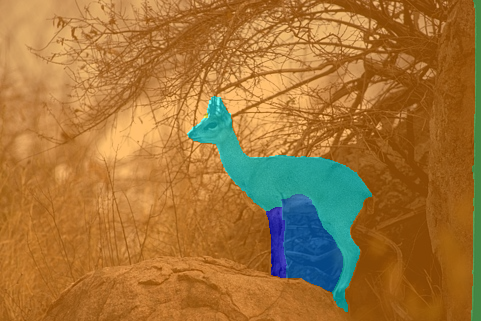}
		
&		\includegraphics[width=0.33\textwidth]{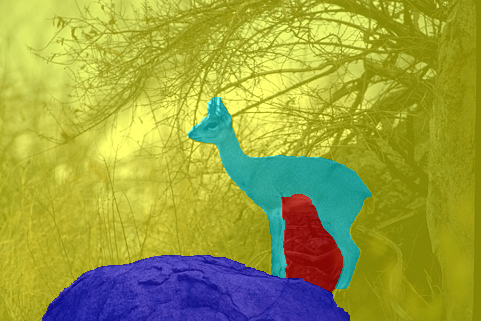} \\         \hline
	    Baseline RCF & Baseline CRF & Adaptive CRF\\
	    \hline
		
	\end{tabular}
	\label{example-fig}
\end{figure}

\begin{figure}[h!]
	\centering
	
	\begin{tabular}{c@{ }c@{ }c}
	    \multicolumn{3}{c}{\includegraphics[width=0.33\textwidth]{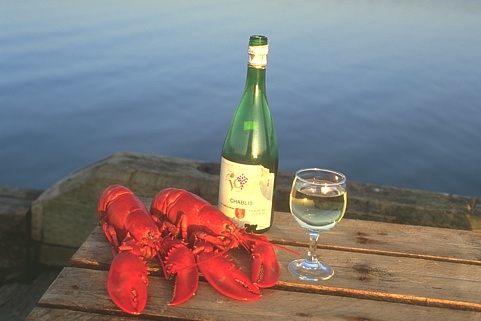}}\\
	    
	    \hline
	    \multicolumn{3}{c}{Edge maps}\\
	    \hline
	    
	    \includegraphics[width=0.33\textwidth]{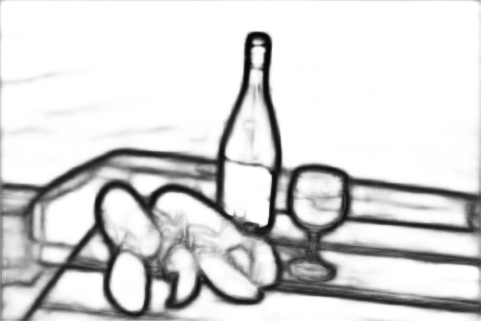}
	     &\includegraphics[width=0.33\textwidth]{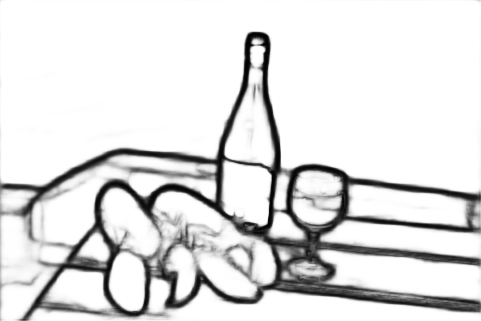}
		
&		\includegraphics[width=0.33\textwidth]{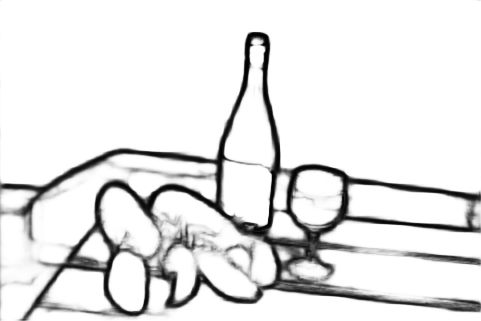} \\
\hline
	    \multicolumn{3}{c}{UCM Segmentations (Threshold=0.5)}\\
	    \hline
		\includegraphics[width=0.33\textwidth]{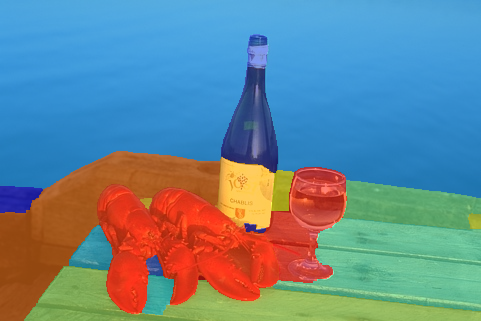}
	     &\includegraphics[width=0.33\textwidth]{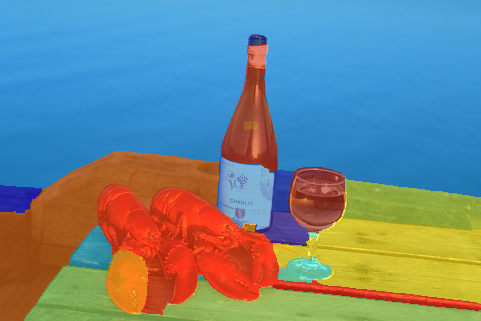}
		
&		\includegraphics[width=0.33\textwidth]{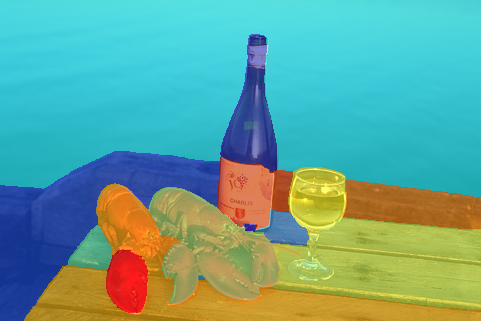} \\         \hline
	    Baseline RCF & Baseline CRF & Adaptive CRF\\
	    \hline
		
	\end{tabular}
	\label{example-fig}
\end{figure}

\end{document}